\documentclass[11pt,letterpaper]{article}
\usepackage{ijcnlp2017}
\usepackage{times}
\usepackage{latexsym}
\usepackage{graphicx}
\usepackage{CJK}
\usepackage{enumerate}
\usepackage{amssymb}
\usepackage{amsfonts}
\usepackage[noend]{algpseudocode}
\usepackage{algorithmicx,algorithm}
\usepackage{booktabs}
\usepackage{latexsym} 
\usepackage{amsmath}
\usepackage{multirow}
\usepackage{url}
\ijcnlpfinalcopy

\title{Towards Neural Machine Translation with Partially Aligned Corpora}

\author{Yining Wang$^{\dagger\ddagger}$, Yang Zhao$^{\dagger\ddagger}$, Jiajun Zhang$^{\dagger\ddagger}$, Chengqing Zong$^{\dagger\ddagger*}$ \and Zhengshan Xue$^\#$\\
 	$^\dagger$National Laboratory of Pattern Recognition, CASIA, Beijing, China \\ 
 	$^\ddagger$University of Chinese Academy of Sciences, Beijing, China \\
	$^*$CAS Center for Excellence in Brain Science and Intelligence Technology, Shanghai, China  \\
	$^\#$Toshiba (China) Co.,Ltd. \\
{\tt \{yining.wang, yang.zhao, jjzhang, cqzong\}@nlpr.ia.ac.cn} \\ 
{\tt xuezhengshan2@toshiba.com.cn}}

\date{}

\begin{document}
\begin{CJK*}{UTF8}{gbsn}
\maketitle

\begin{abstract}
While neural machine translation (NMT) has become the new paradigm, the parameter optimization requires large-scale parallel data which is scarce in many domains and language pairs. In this paper, we address a new translation scenario in which there only exists monolingual corpora and phrase pairs. We propose a new method towards translation with partially aligned sentence pairs which are derived from the phrase pairs and monolingual corpora. To make full use of the partially aligned corpora, we adapt the conventional NMT training method in two aspects. On one hand, different generation strategies are designed for aligned and unaligned target words. On the other hand, a different objective function is designed to model the partially aligned parts. The experiments demonstrate that our method can achieve a relatively good result in such a translation scenario, and tiny bitexts can boost translation quality to a large extent.
\end{abstract}

\section{Introduction}

Neural machine translation (NMT) proposed by Kalchbrenner et al.\shortcite{kalchbrenner2013recurrent}, Sutskever et al.\shortcite{sutskever2014sequence} and Cho et al.\shortcite{cholearning} has achieved significant progress in recent years. Different from traditional statistical machine translation(SMT) \cite{Koehn:2003,Chiang:2005,Liu:2006,zhai2012tree} which contains multiple separately tuned components, NMT builds an end-to-end framework to model the whole translation process. For several language pairs, NMT is reaching  significantly better translation performance than SMT \cite{luongaddressing,wu2016google}. 

In general, in order to obtain an NMT model of great translation quality, we usually need large-scale parallel data. Unfortunately, the large-scale parallel data is always insufficient in many domains and language pairs. Without sufficient parallel sentence pairs, NMT tends to learn poor estimates on low-count events. 


\begin{figure}[!t]
	\centering
	\includegraphics[width=1.0 \columnwidth]{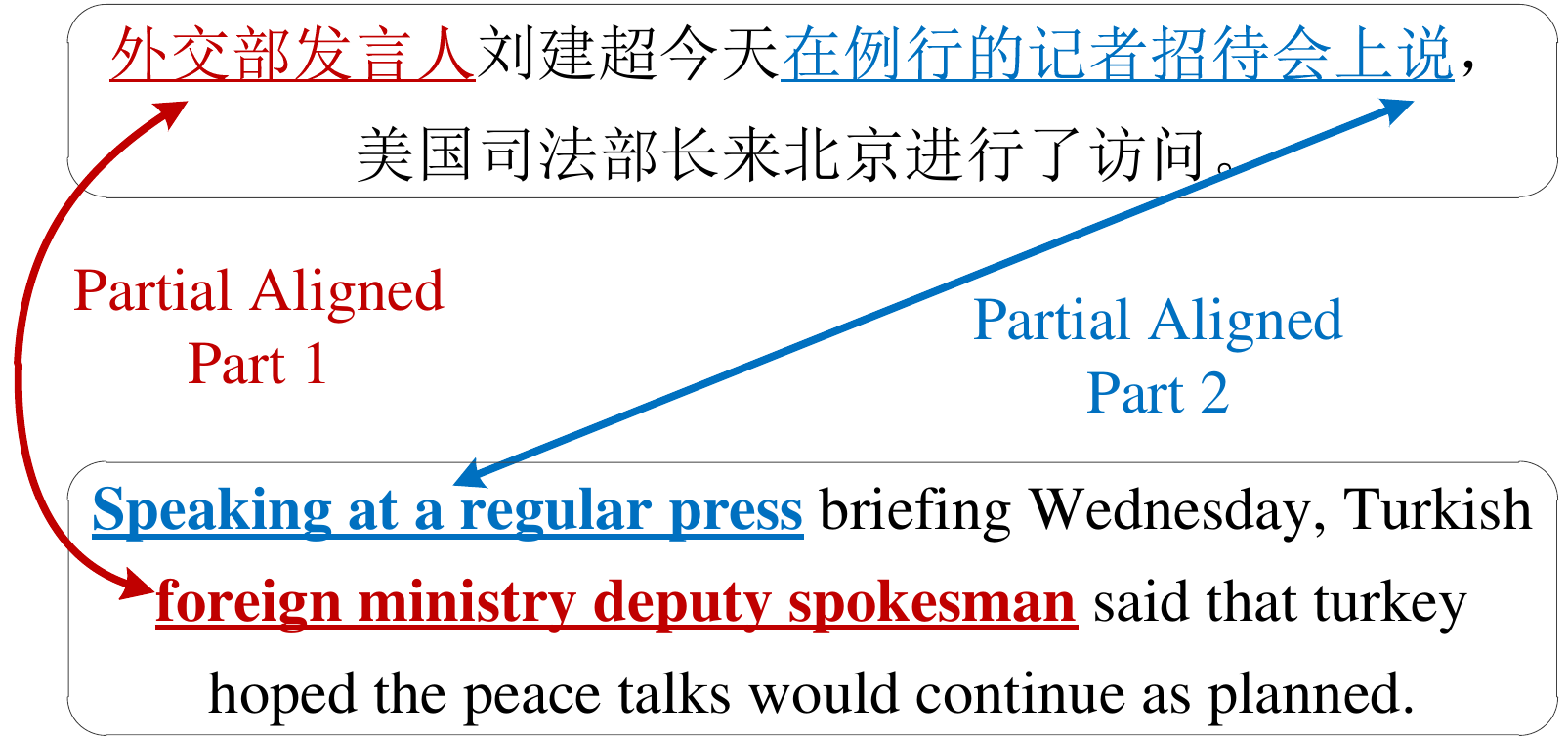}
	\caption{An example of our partially aligned training data, in which the source sentence and target sentence are not parallel but they include two parallel parts (highlight in blue and red respectively).  }
	\label{overview}
\end{figure}

Actually, there have been some effective methods to deal with the situation of translating language pairs with limited resource under different scenarios \cite{Johnson2016Google,chengjoint,sennrich2015improving,zhang2016exploiting}. In this paper, we address a new translation scenario in which we do not have any parallel sentences but have massive monolingual corpora and phrase pairs. The previous methods are hard to be used to learn an NMT model under this situation. In this paper, we propose a novel method to learn an NMT model using only monolingual data and phrase pairs. 

Our main idea is that although there does not exist the parallel sentences, we can derive the sentence pairs which are non-parallel but contain the parallel parts (in this paper, we call these sentences as \textbf{partially aligned sentences}) with the monolingual data and phrase pairs. Then we can utilize these partially aligned sentences to train an NMT model. Figure~\ref{overview} shows an example of our data. Source sentence and target sentence are not fully aligned but contain two translation fragments: ("外交部发言人", "foreign ministry deputy") and ("在例行的记者招待会上说", "speaking at a regular press"). Intuitively, these kinds of sentence pairs are useful in building an NMT model.

To use these partially aligned sentences, the training method should be different from the original methods which are designed for parallel corpora. In this work, we adapt the conventional NMT training method mainly from two  perspectives. On one hand, different generation strategies are designed for aligned and unaligned target words. For aligned words, our method guides the translation process based on both the context of source side and previously predicted words. When generating the unaligned target words , our model only depends on the words previously generated without considering the context of source side. On the other hand, we redesign the objective function so as to emphasize the partially aligned parts in addition to maximizing the log-likelihood of the target sentence.

The contributions of our paper are twofold:

\begin{enumerate}[1)]
    \item Our approach addresses a new translation scenario, where there only exists monolingual data and phrase pairs. We propose a method to train an NMT model under this scenario. The method is simple and easy to implement, which can be used in arbitrary attention-based NMT framework.
    \item  Empirical experiments on the Chinese-English translation tasks under this scenario show that our method can achieve a relatively good result. Moreover, if we only add a tiny parallel corpus, the method can obtain significant improvements in terms of translation quality.
\end{enumerate}

\section{Review of Neural Machine Translation}

Our approach can be easily applied to any end-to-end attention-based NMT framework. In this work, we follow the neural machine translation architecture by Bahdanau et al.~\shortcite{bahdanau2014neural}, which we will summarize in this section.

Given the source sentence $X=\left \{x_{1},x_{2},...,x_{Tx}\right\}$ and the target sentence $Y=\left \{y_{1},y_{2},...,y_{Ty}\right\}$. The goal of machine translation is to transform source sentence into the target sentence. The end-to-end NMT framework consists of two recurrent neural networks, which are respectively called encoder and decoder. First, the encoder network encodes the $X$ into context vectors $C$. Then, the decoder network generates the target translation sentences one word each time based on the context vectors $C$ and target words previously generated. More specifically, that is $p(y_{i}|y_{<i},C)$.

In encoding stage, it transforms $X$ into a sequence of vectors $h_{enc}=\left \{{h_{1}^{k},h_{2}^{k},h_{3}^{k},...,h_{T}^{k}}\right \}$ using $m$ stacked LSTM \cite{hochreiter1997long} layers. Finally, the encoder chooses the hidden states of the top encoder layer as $h_{top}=\left \{h_{1}^{m},h_{2}^{m},h_{3}^{m},...,h_{T}^{m}\right \}$ that we will use in attention mechanism to calculate context vector later.

In decoding stage, it generates one target word at a time from conditional probability $p(y_{i}|y_{<i},C;\theta)$ also via $m$ stacked LSTM layers parameterized by $\theta$. Supposing we have obtained the context vector, the conditional probability $p(y_{i}|y_{<i},C;\theta)$ is calculated as follows:

\begin{equation}
\begin{aligned}
p(y_{i}|y_{<i},C;\theta)&=p(y_{i}|y_{<i},c_{i})\\&= softmax(g(W_{y_{i}},z_{i}^{m},c_{i}))
\end{aligned}
\end{equation}
Where $W_{y_{i}}$ is embedding of the target word, $z_{i}^{m}$ is current hidden states of top layer in decoder network. Note that the first hidden states of decoder $z_{0}^{k}$ are set to the last hidden states of encoder as follows:
\begin{equation}
z_{0}^{k}=h_{T}^{k} 
\end{equation}
$c_{i}$ can be computed as a weighted sum of the source-side $h_{s}$ as follows:
\begin{equation}
c_{i}=\sum_{j=1}^{T_{x}}a_{ij}h_{i}^{m}
\end{equation}

Where $a_{ij}$ is alignment probability, which can be calculated in multiple ways \cite{luongeffective}. In our method, we use a simple single-layer feed forward network. This alignment probability measures how relevant \emph{i}-th context vector of source sentence is in deciding the current symbol in translation. The probability will be further normalized:
\begin{equation}
a_{ij}=\frac{exp(e_{ij})}{\sum_{k=1}^{Tx}exp(e_{ik}))}
\end{equation}

A detailed introduction of the encoder-decoder framework is described in Bahdanau et al.~\shortcite{bahdanau2014neural}. In order to train NMT system, we use parallel data to optimize the network parameters by maximizing the conditional log-likelihood:

\begin{equation}
L(\theta ,D)=\frac{1}{N}\sum_{n=1}^{N}\sum_{i}^{T_{y}}logp(y_{i}^{(n)}|y_{<i}^{(n)},X^{(n)},\theta)
\end{equation}

\section{NMT with Partially Aligned Data}

In \S 2 we gave a brief description of the attention-based NMT models whose network parameters are trained using parallel sentence pairs. However, in the translation scenario where there only exists monolingual corpora and phrase pairs, the conventional NMT framework is hard to be used in training a model. In this section, we first explain how we actually obtain the partially aligned corpora with aligned positions using phrase pairs and monolingual corpora, then introduce our method to train the NMT models using the partially aligned sentences according to the particular properties of the corpora.

\subsection{Constructing partially Aligned Corpora}
Assuming there exists abundant phrase pairs and monolingual sentences in source and target languages, we define our approach to extract partially aligned sentence pairs for training.

Given a phrase pair (ph\_s, ph\_t), ph\_s may appear in a source-side monolingual sentence X, and ph\_t may appear in a target-side monolingual sentence Y. Then, X and Y are non-parallel but contain the parallel part. We call these kinds of data the partially aligned sentences. In this work, we collect the partially aligned sentences by searching the phrase pairs in both of the source and target monolingual data simultaneously. In order to reduce the time of the searching process, the monolingual training corpora are first split into many parts. Then, we retrieve the source phrase in each part to restrict the source range of partially aligned corpus. With the retrieved results, we can search the final results of the partially aligned sentences pairs easily. In this way we can construct our corpora, in which only one or more phrases are aligned in every sentence pairs. We denote a partially aligned sentence ($X=x_{1},x_{2},...x_{Tx}$,$Y=y_{1},y_{2},...,y_{Ty}$), in which a set of the phrase pairs aligned with each other. We call these pairs partially aligned part:

\begin{equation}
\begin{aligned}
&P_{x}^{(k)}=x_{k1},...,x_{kp}\\&P_{y}^{(k)}=y_{k1},...,x_{kq}
\end{aligned}
\end{equation}
$P_{x}^{(k)}$ and $P_{y}^{(k)}$ are the phrases in the source and target sentences respectively, and they are translation equivalents.




\subsection{Model Descriptions}
In \S 3.1, we acquired the partially aligned corpora with the phrase pairs and monolingual sentences. Now, we need to use them to train the NMT model. As the traditional NMT model is designed for the parallel sentences, it is not suitable for partially aligned sentences. Thus we redesign the traditional NMT model as follows. Figure~\ref{frame} shows the basic framework of our training method. Our model has 4 different parts from conventional NMT model, including initial states, generation process, objective functions and vocabulary size.
\subsubsection{Initial States}
\begin{figure}[!t]
	\centering
	\includegraphics[width=1.0 \columnwidth]{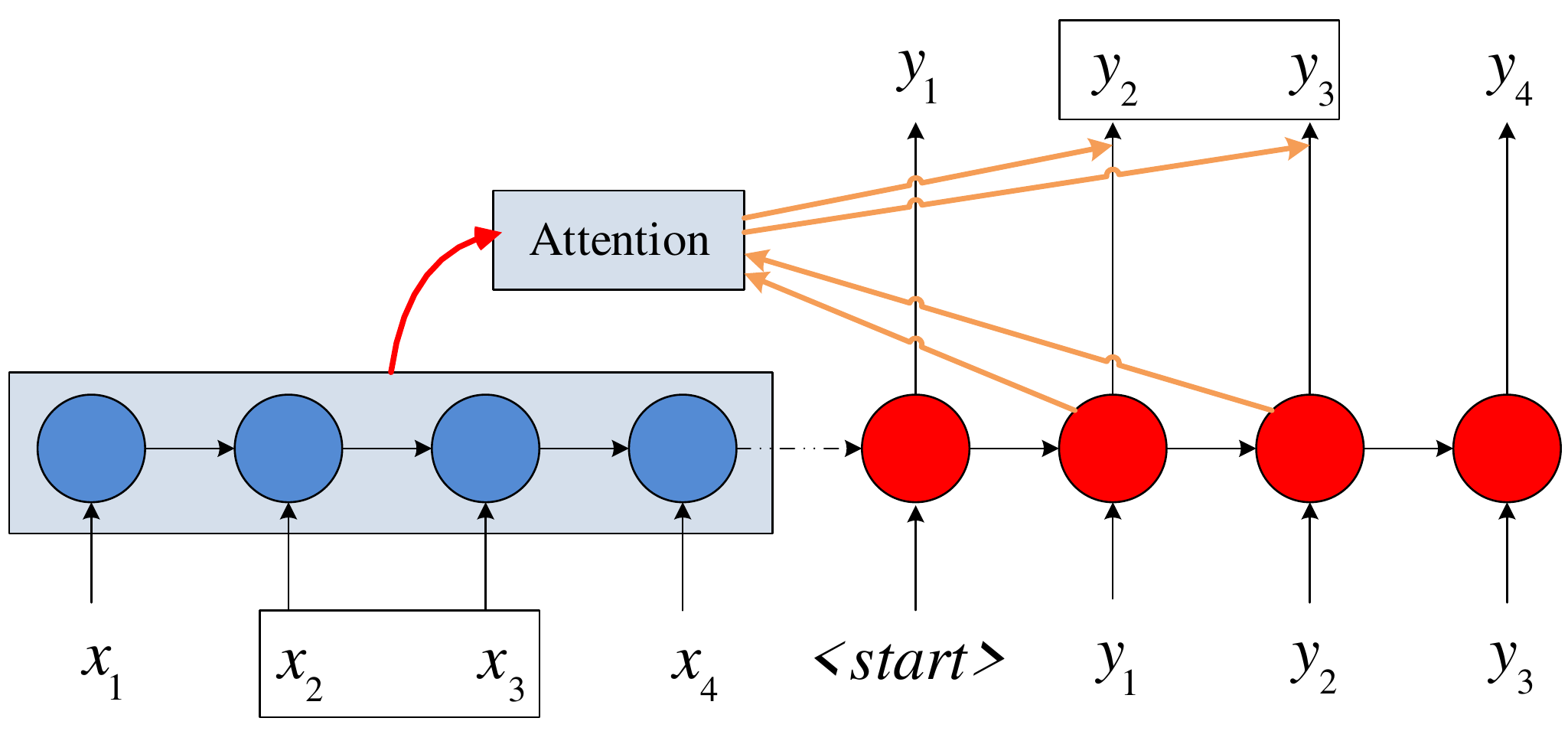}
	\caption{The framework of our training method for partially aligned sentence, in which ($x_{2}$,$x_{3}$) and ($y_{2}$,$y_{3}$) are parallel parts.  }
	\label{frame}
\end{figure}

The first difference is the initial hidden states of decoder. In the conventional NMT model, the initial hidden states of decoder $z_{0}^{k}$ is set to the last hidden state in encoder $h_{Tx}^{k}$, as shown in Eq. (2).  For parallel sentences, this setting is reasonable, while for partially aligned sentence, this initial method is inappropriate. The reason is that the hidden state of last word in source sentence is irrelevant to the target sentence, considering the fact that under our scenario, the target side sentences may entirely uncorrelated to the source sentences. Thus, in our model, $z_{0}^{k}$ is set to a zero vector as follows:
\begin{equation}
 Z_{0}^{k}=\textbf{0}
 \end{equation}
In Figure~\ref{frame}, the hidden state of "$<$start$>$" symbol is set to zero vector when $y_{1}$ does not belong to parallel part of the sentence pair.

\subsubsection{Generation Process}
The second difference is the generation process. In the conventional NMT system, the model generates each target word based on the context vector $c_{i}$ and previously predicted words $y_{<i}$ as shown in Eq. (1). This generation strategy is unsuitable for the partially aligned corpora, since there exists many unaligned target words. Intuitively, when the model generates the non-parallel parts, it is unnecessary to take the context vector $c_{i}$ into consideration. As Figure~\ref{frame} illustrated, ($x_{2}$,$x_{3}$) and ($y_{2}$,$y_{3}$) are parallel parts in this partially aligned sentence pair, and we use context vector which is generated by attention mechanism only when the decoder outputs $y_{2}$ and $y_{3}$. Therefore, the decoder in our model can be described as follows:  
\begin{equation}
c_{i}=\left\{\begin{matrix}
\sum_{j}^{Tx}a_{ij}h_{i} \ \ \  if \ y_{i} \in P_{y}^{(k)} \\ 
\ \  \ \ \  \ 0 \ \ \ \ \ \ \ \ \ \  \ if \ y_{i} \notin  P_{y}^{(k)}
\end{matrix}\right.
\end{equation}
where $a_{ij}$ is calculated as Eq. (2), $P_{y}^{(k)}$ is the target partial part, as shown in Eq. (5). In Eq. (8), our model generates the aligned target words based on the context vector $c_{i}$ and previously predicted words $y_{<i}$. When generating the unaligned target words, the model sets the context vector $c_{i}$ to zero, indicating that the model generates these words only based on the LSTM-based RNN language model.

\subsubsection{Objective Function}
Third, we redesign the objective function. Given the parallel data, the objective function is to maximize the log-likelihood of each target word as shown in Eq. (5). For the partially aligned sentence, besides the source and target sentence, we know the phrase alignment information. Hence, apart from maximizing the log-likelihood of each target word, we also hope to make the source and target words in partially aligned part align to each other. As shown in Figure~\ref{frame}, when predicting the words $y_{2}$ and $y_{3}$, we want to attend more information of corresponding words $x_{2}$ and $x_{3}$. Thus, we inject an auxiliary object function to achieve it. More specifically, our objective function is designed as follows:
\begin{equation}
\begin{aligned}
 L_{P} &(\theta , D)= \\& \frac{1}{N}\sum_{n=1}^{N} \{\sum_{i}^{T_{y}}log(p(y_{i}^{(n)}|y_{<i}^{(n)},X^{(n)},\theta)\\& +\lambda
\times \triangle (a^{(n)},\widehat{a}^{(n)},\theta))\} 
\end{aligned}
\end{equation}
Where $a^{(n)}$ is defined in Eq. (4), $\triangle$ is a loss function that encourages the agreement between  $a^{(n)}$ and
$\widehat{a}^{(n)}$. $\widehat{a}^{(n)}$ is the supervised attention determined by alignment relationship between $P_{X}$ and $P_{Y}$, and can be calculated as follows:
\begin{equation}
\widehat{a}_{i,j}^{(n)}=\left \{ \begin{matrix}
1\ \ \ \ if \ x_{j}\in P_{X} \ and \ y_{i}\in P_{Y}  \\ 
0 \ \ \ \ \ \ \ \ \ \ \ \ \ \ \ others \ \ \ \ \ \ \ \ \ \ \ \ \ \  \ \ \ \
\end{matrix} \right.
\end{equation}
$\lambda > 0 $ is a hyper-parameter that balances the preference between likelihood and agreement. In this paper, it is set to 0.3.

As shown in Eq. (8), our objective function does not only consider to maximize the log-likelihood of the target sentence, but also encourages the alignment $a_{ij}$ produced by NMT to have a larger agreement with the prior alignment information. This objective function is similar to that used by the supervised attention method~\cite{mi2016supervised,Liu2016superattention}. Inspired by Liu et al.~\shortcite{Liu2016superattention}, the agreement between $a^{(n)}$ and $\widehat{a}^{(n)}$ can be defined in different ways:

\begin{itemize}
\item Multiplication (MUL)
\begin{equation}
\begin{aligned}
 &\triangle (a^{(n)},\widehat{a}_{i,j}^{(n)};\theta)\\&=-\sum_{i=0}^{Ty}\sum_{j=0}^{Tx}a(\theta)_{i,j}^{(n)}\times \widehat{a}_{i,j}^{(n)}
 \end{aligned}
\end{equation}
where $\widehat{a}_{i,j}^{(n)}$ is computed by Eq. (10)

\item Mean Squared Error (MSE)

\begin{equation}
\begin{aligned}
 &\triangle (a^{(n)},\widehat{a}_{i,j}^{(n)};\theta)\\&=-\sum_{i=0}^{Ty}\sum_{j=0}^{Tx}\frac{1}{2}(a(\theta)_{i,j}^{(n)}-\widehat{a}_{i,j}^{(n)})^{2}
 \end{aligned}
\end{equation}

\end{itemize}

\subsubsection{Limited Vocabulary}
The last difference is the vocabulary size during decoding. To make use of phrase pairs as much as possible, we extract a number of special phrase pairs whose source and target are both one word. In decoding phase, as what Mi et al.~\shortcite{Mi2016Vocabulary} have done, we can use these special phrase pairs to reduce the vocabulary size when computing the final score distribution. In this way, we can not only acquire more accurate translation of each word, but also accelerate the decoding speed. The vocabulary size can be reduced as follows:

\begin{equation}
V=V_{1} \cup V_{2}
\end{equation}
Where $V_{1}$ contains the most frequently target words and $V_{2}$ is a target words set. This set $V_{2}$ is made up of all the target words of the special phrase pairs whose corresponding source words belong to the source sentence.

\section{Experiment}
In this section, we perform the experiment on Chinese-English translation tasks to test our method. 
\subsection{Dataset}
We evaluate our approach on large-scale monolingual data set from LDC corpus, which includes 13M Chinese sentences and 10M English sentences. Table 1 shows the detailed statistics of our training data. To test our model, we use NIST 2003(MT03) as development set, and NIST 2004-2006(MT04-06) as test set. The evaluation metric is case-insensitive BLEU \cite{papineni2002bleu} as calculated by the $multi\text{-}bleu.perl$.

\begin{table}[htbp]

\begin{tabular}{|c|c|c|c|}
\hline
Corpus & & Chinese & English\\
 \hline
 \hline
 \multirow{3}{*}{monolingual}& \#Sent. & 13.33M & 10.03M \\
 \cline{2-4}
 &\#Word&327.10M & 276.07M \\
  \cline{2-4}
 &Vocab & 1.83M & 1.07M \\
  \hline
 \end{tabular}
\caption{\label{tab:test} The statistics of monolingual dataset on the LDC corpus.}
\end{table}
\begin{table*}[htbp]
\centering
 \begin{tabular}{ccccccc}
  \toprule
  \# & \textbf{System} & \textbf{MT03} & \textbf{MT04} & \textbf{MT05} & \textbf{MT06} & \textbf{Ave} \\
  \midrule
  \midrule
1 & Phrase NMT Model    & 3.64 & 4.25 & 3.55 & 3.77 & 3.80 \\
2 & Partially Aligned Model(MUL) & 3.80 & 4.37 & 3.75 & 4.24 & 4.04 \\
3 & Partially Aligned Model(MSE) &  5.11 & 5.04 & 4.26 & 4.95 & 4.84  \\
4 & Partially Aligned Model(MSE) + LimitedVocab  & \textbf{6.63} & \textbf{6.81} & \textbf{5.59} & \textbf{5.77} & \textbf{6.20} \\
5 & Phrase NMT model + LimitedVocab & 3.78 & 4.33 & 3.63	& 3.94	& 3.92\\
  \bottomrule
 \end{tabular}
\caption{\label{tab:test} Translation results (BLEU score) for different translation methods.}
\end{table*}
\subsection{Data Preparing and Preprocessing}
Considering the fact that the amount of manually annotated phrase pairs is not enough, in order to imitate the environment of experiment, we extract phrase pairs from parallel corpora automatically to make up for the shortage of quantity. To do this, we use Moses \cite{koehn2007moses} in its training step to learn a phrase table from LDC corpus, which includes 0.63M sentence pairs. In order to simulate the experiment as far as possible, we adopt three strategies to filter low quality phrase pairs: 1) the phrases containing the punctuation should be filtered out. (The special phrase parirs introduced in \S 3.2.4 should be retain) 2) the length of source phrase and target phrase should be greater than 3. 3) only the phrase pairs whose translation probability exceed 0.5 should be retain. In this way, we can get 3M phrase pairs in our experiment. According to our analysis, the average length of phrases are 4.15 and 4.70 on source and target side respectively.

When we search the phrase pairs in monolingual sentences, an obstacle is that one phrase pair will get different source sentences with same target sentence or same source sentence with different target sentences. Therefore, for one phrase pair, we have to restrict the number $n$ of both source sentences and target sentences. To balance the search speed of the phrase pairs in monolingual corpora and the amount of partially aligned sentences, we set the hyper-parameter $n$ to 7. We can search for 5M partially aligned sentences in our experiment. We also calculate the average length ratio of aligned phrases against the whole sentence, which is only 21\% and 23\% respectively on source and target side.

To ensure the quality of the partially aligned corpora, we also set the number of phrases that aligned to each other in one sentence pair must be greater than a threshold. Here, the threshold is set to 2. That is to say the partially aligned sentence pair should contain at least two aligned phrase pairs.


\subsection{Training Details}
We build our described method based on the Zoph\_RNN toolkit\footnote{\url{https://github.com/isi-nlp/Zoph RNN}} written in C++/CUDA. Both encoder and decoder consist of two stacked LSTM layers. We set minibatch size to 128. The word embedding dimension of both source and target sides is 1000, and the dimensions of hidden layers unit is set to 1000. In our baseline model, we limit the vocabulary of both source and target languages to 30K most frequent words, and other words are replaced by a special symbol “UNK”. We run our model on the training corpus 20 iterations in total with stochastic gradient decent algorithm. We set learning rate to 0.1 at the beginning and halve the threshold while the perplexity increases on the development set. Dropout is applied to our model, and the rate is set to 0.2. For testing, we employ beam search algorithm, and the beam size is 12.

\subsection{Training Methods}
We conduct our experiment on the dataset mentioned above, and we list the training methods used as follows:
\begin{enumerate}[1)]
\item \textbf{Phrase NMT Model:} As mentioned above, the only parallel resource is phrase pairs. We use attention-based NMT system to train only on the 3M phrase pairs to get our baseline result.
\item \textbf{Partially Aligned Model(MUL):} We train our NMT model using the objective function of multiplication method on the partially aligned sentences.
\item \textbf{Partially Aligned Model(MSE):} We train our NMT model using the objective function of Mean Squared Error(MSE) method.
\item \textbf{Partially Aligned Model(MSE) + LimitedVocab:} It is similar to Partially Aligned Model(MSE) and the only difference is that we restrict the final score distribution on a limited target vocabulary, which is described in \S 3.2.4.
\item \textbf{Phrase NMT Model + LimitedVocab:} It is the method that LimitedVocab is used in Phrase NMT model.
\end{enumerate}

\section{Results and Analysis}
\label{sec:blind}
\subsection{Phrase NMT Model vs. Partially Aligned Method}

We present the translation results in BLEU scores of different systems in Table 2. Our first concern is whether the proposed model can actually improve the translation quality. As Table 2 shows, we find that our partially aligned model (both MUL supervised method and MSE supervised method) is superior to the Phrase NMT Model, which indicates that our Partially Aligned method is effective in improving the translation quality.

\begin{figure}[h]
	\includegraphics[width=1.0 \columnwidth]{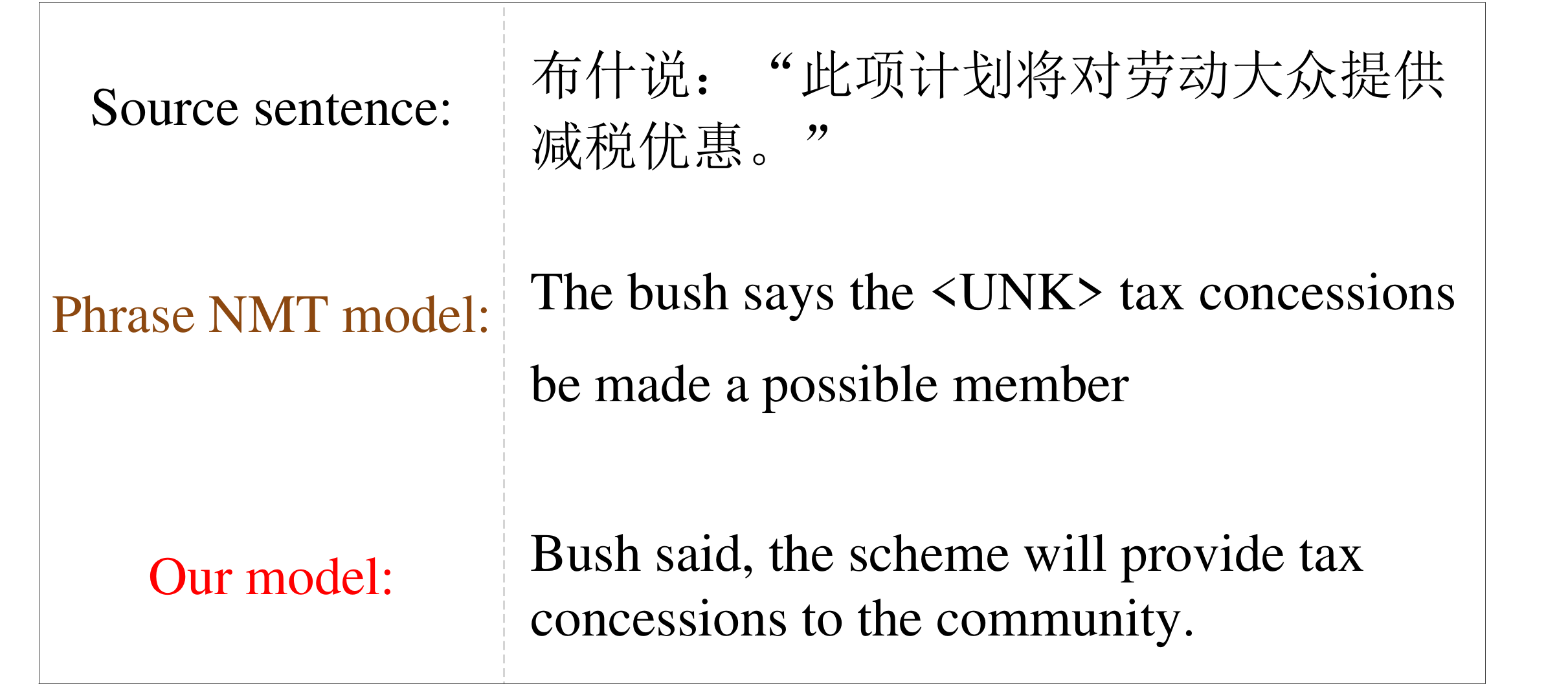}
	\caption{A translation example comparing between phrase NMT model and our partially aligned method. }
	\label{fig2}
\end{figure} 

Figure~\ref{fig2} lists a comparison example of Phrase NMT model and our model. Obviously, our model can achieve the correct translation while the Phrase NMT model generates the unfaithful result. It demonstrates that our model is actually having the ability to learn translation adequacy from aligned parts and fluency from both aligned and unaligned parts.

In \S 3.2, we described two objective functions in our model. We focus on the difference of these two approaches. As a comparison, MSE method (Row 3) outperforms the MUL method (Row 2) with an average improvement of 0.8 BLEU points, indicating that MSE method is more effective as an object function for our partially aligned model. In the following experiment, we adopt the MSE method as the new objective function used in our model.

\subsection{Effect of Limited Vocabulary}
In Table 2, an interesting result is that using a reduced vocabulary can significantly improve the performance (+1.36 BLEU points), but it can only achieve 0.12 BLEU points improvement for Phrase NMT model. According to Mi et al.~\shortcite{Mi2016Vocabulary}, this approach is useful in conventional NMT model. Our result is in agreement with their findings, and the improvement is more prominent in our partially aligned model. Under our scenario, compared to the parallel corpora, fewer parallel parts appear in sentence pairs. The faithfulness of our translation result is relatively poor while the fluency is relatively good. With the limited relevant vocabulary, the faithfulness of the translation results is much improved.


\begin{figure}[h]
	\includegraphics[width=1.0 \columnwidth]{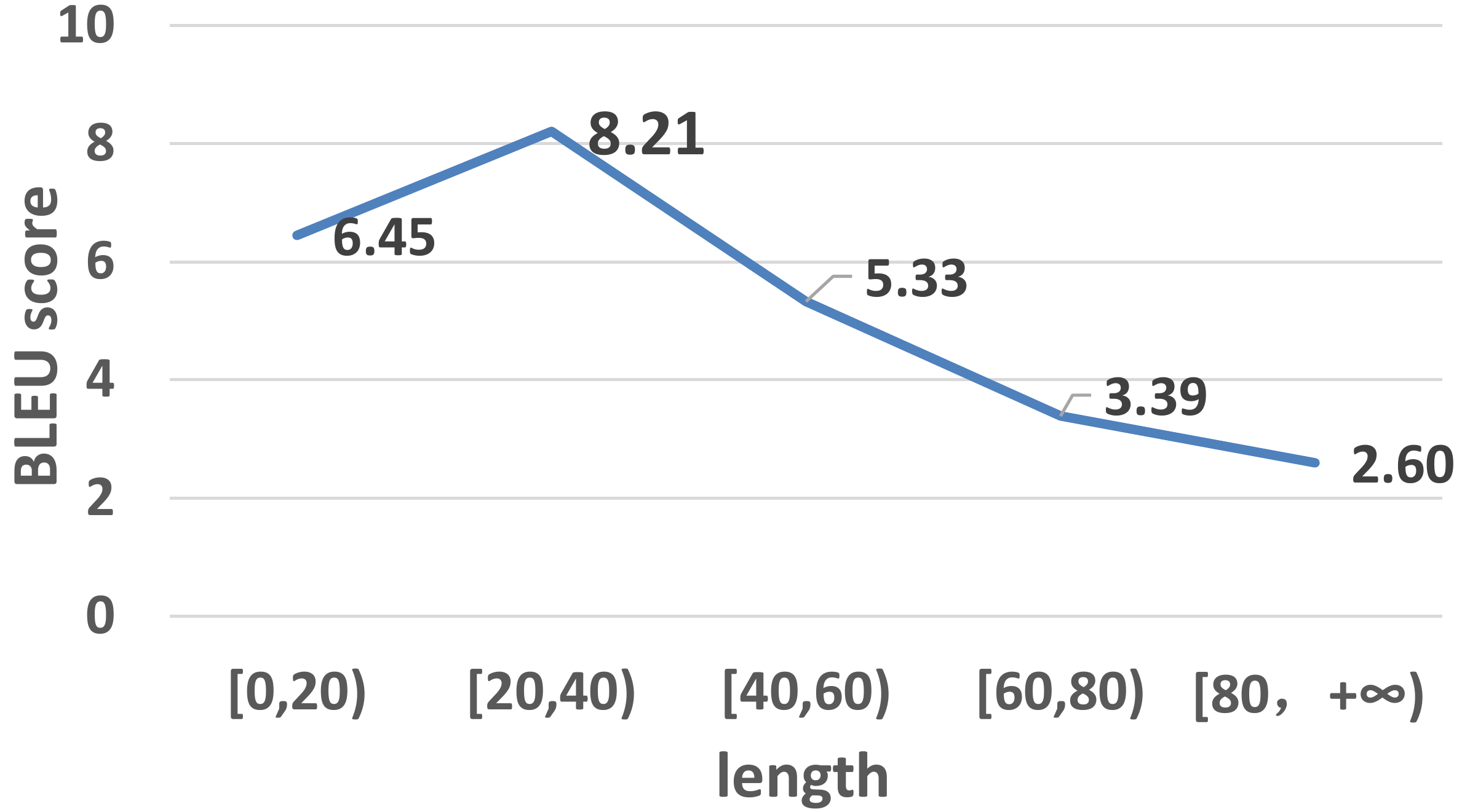}
	\caption{Translation results (BLEU score) of different lengths.}
	\label{fig}
\end{figure}

\subsection{Result of Different Sentence Lengths}

The performance of our partially aligned model with different lengths is another problem we care about. We randomly select 1000 sentences from translation results of test set (MT03-08), which are trained by the Partially Aligned Model(MSE)+ReduceDict method. We classify them into five categories according to the length. Figure~\ref{fig} shows the results of the experiment.


We find that the sentences with shorter length (lower than 40) yield better results than the long sentences. When the length of sentences exceeds 80, the quality of translation is rather poor. The reason is that we mainly use phrase pairs in our method, and the length of phrase is relative short compared to whole sentence. So our model is more suitable for the translation of short sentences. When we translate long sentences, the parameters in our model are not adjusted for tuning, and our approach can not produce translation of high quality.

\subsection{Effect of Adding Small Parallel Corpus}
We concern that when we have a tiny parallel corpus, whether the small scale parallel corpus can boost the translation performance of the partially aligned method. Here, we fine-tune the partially aligned translation model on these corpora. The details of these corpora are introduced in Table 3.

\begin{table}[htbp]
\centering
\begin{tabular}{|c|c|c|c|}
\hline
Corpus & & Chinese & English\\
 \hline
  \hline
 \multirow{3}{*}{parallel}& \#Sent. &\multicolumn{2}{|c|}{0.1M} \\
  \cline{2-4}
 &\#Word & 3.00M & 3.86M \\
  \cline{2-4}
 & Vocab & 0.07M & 0.04M \\
 \hline
 \end{tabular}
\caption{\label{tab:test} The statistics of small-scale parallel datasets.}
\end{table}

\begin{table*}[htbp]
\centering
 \begin{tabular}{ccccccc}
  \toprule
   \textbf{\#Sent.} & \textbf{Method} & \textbf{MT03} & \textbf{MT04} & \textbf{MT05} & \textbf{MT06}  & \textbf{Ave} \\
  \midrule
  \midrule
   \multirow{2}{*}{20K} & \textbf{NMT Model} & 1.60 & 1.22 & 1.05 & 1.70 & 1.39\\
    & \textbf{Partially Aligned Model(MSE) + Para} & 12.36 & 15.07 & 11.64 & 14.61 & 13.42 \\
  \midrule    
   \multirow{2}{*}{40K} & \textbf{NMT Model} & 1.87 & 2.00 & 1.47 & 2.24 & 1.90\\
    & \textbf{Partially Aligned Model(MSE) + Para} & 14.12 & 17.84 & 13.66 & 17.26 & 15.72\\
  \midrule    
   \multirow{2}{*}{60K} & \textbf{NMT Model} & 3.72 & 5.04 & 3.49 & 4.47 & 4.18 \\
    & \textbf{Partially Aligned Model(MSE) + Para} & 15.54 & 19.62 & 15.16 & 19.25 & 17.39\\
  \midrule
   \multirow{2}{*}{80K} & \textbf{NMT Model} & 7.96 & 11.85 & 8.16 & 10.53 & 9.63 \\
    & \textbf{Partially Aligned Model(MSE) + Para} & 17.16 & 21.18 & 16.65 & 20.61 & 18.90\\
  \midrule
   \multirow{2}{*}{100K} & \textbf{NMT Model} & 14.50 & 18.21 & 14.29 & 17.49 & 16.12\\
    & \textbf{Partially Aligned(MSE) Model + Para} & 18.30 & 22.50 & 17.92 & 21.55 & 20.07 \\
  \bottomrule
 \end{tabular}
\caption{\label{tab:test} Effect of different data size of parallel corpus. Method NMT Model means the result of conventional NMT system trained on these low-count parallel sentences. Partially Aligned Model(MSE) + Para means the result of our model fine-tuned by these parallel sentences. }
\end{table*}

The result is presented in Table 4. We can observe that the translation result tends to be poor by only using a small-scale parallel corpora. It indicates that conventional NMT system cannot learn a good model on the small-scale datasets. However, when fine-tuning our partially aligned model with this small parallel corpus, we can get a surprising improvement. The results suggest that when under a scenario in which we only have monolingual corpora and phrase pairs, even a few bitexts can boost translation quality to a large extent.


We investigate the effect of the different corpora size on the final translation results. According to Table 4, when the number of parallel sentences is quite small (lower than 60K), we can acquire a measurable improvement (more than 10 BLEU) compared to the conventional NMT system result. Especially, when the size of sentence pairs is 40K and 60K, our method obtains the enormous improvement over the NMT model by +13.82 BLEU points and +13.21 BLEU points respectively. When using more than 60K sentence pairs, we still get a relatively high promotion of translation quality. However, the promotion is not very remarkable as Row1-3 reveal in Table 4. We can see when the number of parallel corpora is 100K(Row 5), the improvement over NMT Model is +3.95 BLEU points, which indicates that as the size of parallel corpora increases, the improvement of fine-tuning model is decreasing. 

\section{Related Work}
Most of existing work in neural machine translation focus on integrating SMT strategies \cite{he2016improved,Zhou:2017,wang2017neural,shen2015minimum}, handling rare words \cite{litowards,sennrich2015neural,luongaddressing} and designing the better framework \cite{tu2016coverage,luongeffective,meng2016interactive}. As for translation scenarios, training NMT model under different scenarios has drawn intensive attention in recent years. Actually, there have been some effective methods to deal with them. We divide the related work into three categories:

\subsection{Pivot-based Scenario}
Pivot-based scenario assumes that there only exists source-pivot and pivot-target parallel corpora, which can be used to train source-to-pivot and pivot-to-target translation models. Cheng et al.~\shortcite{chengjoint} propose to translate source language into pivot language, and then the pivot language will be translated into target language. According to the fact that parallel sentences should have close probabilities of generating a sentence in a third language, Chen et al.~\shortcite{chenTS} construct a Teacher-Student framework, in which existing pivot-target NMT model guides the learning process of the source-target model. 

\subsection{Multilingual Scenario}
In multilingual scenario, there exists multiple language pairs but no source-target sentence pairs. Johnson et al.~\shortcite{Johnson2016Google} use parallel corpora of multiple languages to train a universal NMT model. This universal model learns translation knowledge from multiple different languages, which makes zero-shot translation feasible. Firat et al.~\shortcite{Firat2016Zero} present a multi-way, multilingual model to resolve the zero-resource translation. They use other language to train a multi-way NMT model. The model generates pseudo parallel corpora to fine-tune attention mechanism, so as to achieve better translation.

\subsection{Monolingual Data Scenario}
In this scenario, an NMT model of good quality has been trained on existing parallel corpora, but a preferable translation result is still in need by incorporating additional data resource. G{\"u}l{\c{c}}ehre et al.~\shortcite{gulccehre2015using} propose to incorporate target-side corpora as a language model. Sennrich et al.~\shortcite{sennrich2015improving} attempt to enhance the decoder network model of NMT by incorporating the target-side monolingual data. Luong et at.~\shortcite{luongmulti} explore the sequence autoencoders and skip-thought vectors method to exploit the monolingual data of source language. Zhang and Zong~\shortcite{zhang2016exploiting} propose two approaches, self-training algorithm and multi-task learning framework, to incorporate source-side monolingual data. Besides that, Cheng et al.~\shortcite{cheng2016semi} have explored the usage of both source and target monolingual data using a semi-supervised method to reconstruct both source and target side monolingual language, where two NMT frameworks will be used. 

Above methods are designed for different scenarios, and their work can achieve great results on these scenarios. However, when in the scenario we propose in this work, that is we only have monolingual sentences and some phrase pairs, their methods are hard to be utilized to train an NMT model. Under this scenario, monolingual data can be acquired easily, and high quality phrase pairs can be obtained using some effective methods \cite{zhang:2014}. To learn a good NMT model in our translation scenario, we adapt the conventional training procedure by designing a different generation mechanism and a different objective function.

\section{Conclusion}
In this paper, we have presented a new translation scenario for NMT in which we have only monolingual data and bilingual phrase pairs. We obtain large-scale partially aligned sentence pairs from the monolingual data and phrase pairs by an information retrieval algorithm. The generation process and objective function are specially designed in NMT training to take full advantage of the partially aligned corpora. The empirical experiments show that the proposed method is capable to achieve a relatively good result. We further find that only a little amount of parallel sentences can significantly boost the translation quality.

We also notice that the proposed approach with only partially aligned data cannot obtain high translation quality. In the future, we plan to design better approaches to model the partially aligned corpus. We also attempt to evaluate our approach on other language pairs, especially low-resource language pairs.


\section*{Acknowledgments}

The research work has been funded by the Natural Science Foundation of China under Grant No. 61333018 and No. 61402478, and it is also supported by the Strategic Priority Research Program of the CAS under Grant No. XDB02070007

\bibliography{ijcnlp2017}

\begin{thebibliography}{35}
\expandafter\ifx\csname natexlab\endcsname\relax\def\natexlab#1{#1}\fi

\bibitem[{Bahdanau et~al.(2015)Bahdanau, Cho, and Bengio}]{bahdanau2014neural}
Dzmitry Bahdanau, Kyunghyun Cho, and Yoshua Bengio. 2015.
\newblock Neural machine translation by jointly learning to align and
  translate.
\newblock \emph{In Proceedings of ICLR 2015}.

\bibitem[{Chen et~al.(2017)Chen, Cheng, Liu, and Victor}]{chenTS}
Yun Chen, Yong Cheng, Yang Liu, and Li~Victor, O.K. 2017.
\newblock A teacher-student framework for zero-resource neural machine
  translation.
\newblock \emph{In Proceedings of ACL 2017}.

\bibitem[{Cheng et~al.(2017)Cheng, Liu, Yang, Sun, and Xu}]{chengjoint}
Yong Cheng, Yang Liu, Qian Yang, Maosong Sun, and Wei Xu. 2017.
\newblock Joint training for pivot-based neural machine translation.
\newblock \emph{In Proceedings of IJCAI 2017}.

\bibitem[{Cheng et~al.(2016)Cheng, Xu, He, He, Wu, Sun, and
  Liu}]{cheng2016semi}
Yong Cheng, Wei Xu, Zhongjun He, Wei He, Hua Wu, Maosong Sun, and Yang Liu.
  2016.
\newblock Semi-supervised learning for neural machine translation.
\newblock \emph{In Proceedings of ACL 2016}.

\bibitem[{Chiang(2005)}]{Chiang:2005}
David Chiang. 2005.
\newblock A hierarchical phrase-based model for statistical machine
  translation.
\newblock \emph{In Proceedings of ACL 2005}.

\bibitem[{Cho et~al.(2014)Cho, Gulcehre, Bahdanau, Schwenk, and
  Bengio}]{cholearning}
Kyunghyun Cho, Bart van Merri{\"e}nboer~Caglar Gulcehre, Dzmitry Bahdanau,
  Fethi Bougares~Holger Schwenk, and Yoshua Bengio. 2014.
\newblock Learning phrase representations using rnn encoder--decoder for
  statistical machine translation.
\newblock \emph{In Proceedings of EMNLP 2014}.

\bibitem[{Firat et~al.(2016)Firat, Sankaran, Alonaizan, Vural, and
  Cho}]{Firat2016Zero}
Orhan Firat, Baskaran Sankaran, Yaser Alonaizan, Fatos T.~Yarman Vural, and
  Kyunghyun Cho. 2016.
\newblock Zero-resource translation with multi-lingual neural machine
  translation.
\newblock \emph{In Proceedings of EMNLP 2016}.

\bibitem[{G{\"u}l{\c{c}}ehre et~al.(2015)G{\"u}l{\c{c}}ehre, Firat, Xu, Cho,
  Barrault, Lin, Bougares, Schwenk, and Bengio}]{gulccehre2015using}
Caglar G{\"u}l{\c{c}}ehre, Orhan Firat, Kelvin Xu, Kyunghyun Cho, Lo{\i}c
  Barrault, Huei-Chi Lin, Fethi Bougares, Holger Schwenk, and Yoshua Bengio.
  2015.
\newblock On using monolingual corpora in neural machine translation.
\newblock \emph{CoRR, abs/1503.03535}.

\bibitem[{He et~al.(2016)He, He, Wu, and Wang}]{he2016improved}
Wei He, Zhongjun He, Hua Wu, and Haifeng Wang. 2016.
\newblock Improved neural machine translation with smt features.
\newblock \emph{In Proceedings of AAAI 2016}.

\bibitem[{Hochreiter and Schmidhuber(1997)}]{hochreiter1997long}
Sepp Hochreiter and J{\"u}rgen Schmidhuber. 1997.
\newblock Long short-term memory.
\newblock \emph{Neural computation}, 9(8):1735--1780.

\bibitem[{Johnson et~al.(2016)Johnson, Schuster, Le, Krikun, Wu, Chen, Thorat,
  Viégas, Wattenberg, and Corrado}]{Johnson2016Google}
Melvin Johnson, Mike Schuster, Quoc~V Le, Maxim Krikun, Yonghui Wu, Zhifeng
  Chen, Nikhil Thorat, Fernanda Viégas, Martin Wattenberg, and Greg Corrado.
  2016.
\newblock Google's multilingual neural machine translation system: Enabling
  zero-shot translation.
\newblock \emph{arXiv preprint arXiv:1611.04558}.

\bibitem[{Kalchbrenner and Blunsom(2013)}]{kalchbrenner2013recurrent}
Nal Kalchbrenner and Phil Blunsom. 2013.
\newblock Recurrent continuous translation models.
\newblock \emph{In Proceedings of EMNLP 2013}.

\bibitem[{Koehn et~al.(2007)Koehn, Hoang, Birch, Callison-Burch, Federico,
  Bertoldi, Cowan, Shen, Moran, Zens et~al.}]{koehn2007moses}
Philipp Koehn, Hieu Hoang, Alexandra Birch, Chris Callison-Burch, Marcello
  Federico, Nicola Bertoldi, Brooke Cowan, Wade Shen, Christine Moran, Richard
  Zens, et~al. 2007.
\newblock Moses: Open source toolkit for statistical machine translation.
\newblock \emph{In Proceedings of ACL 2007}, pages 177--180.

\bibitem[{Koehn et~al.(2003)Koehn, Och, and Marcu}]{Koehn:2003}
Philipp Koehn, Franz~J. Och, and Daniel Marcu. 2003.
\newblock Statistical phrase-based translation.
\newblock \emph{In Proceedings of ACL-NAACL 2013}.

\bibitem[{Li et~al.(2016)Li, Zhang, and Zong}]{litowards}
Xiaoqing Li, Jiajun Zhang, and Chengqing Zong. 2016.
\newblock Towards zero unknown word in neural machine translation.
\newblock \emph{In Proceedings of IJCAI 2016}.

\bibitem[{Liu et~al.(2016)Liu, Utiyama, Finch, and
  Sumita}]{Liu2016superattention}
Lemao Liu, Masao Utiyama, Andrew Finch, and Eiichiro Sumita. 2016.
\newblock Neural machine translation with supervised attention.
\newblock \emph{In Proceedings of COLING 2016}.

\bibitem[{Liu et~al.(2006)Liu, Liu, and Lin}]{Liu:2006}
Yang Liu, Qun Liu, and Shouxun Lin. 2006.
\newblock Tree-to-string alignment template for statistical machine
  translation.
\newblock \emph{In Proceedings of ACL 2006}.

\bibitem[{Luong et~al.(2016)Luong, Le, Sutskever, Vinyals, and
  Kaiser}]{luongmulti}
Minh-Thang Luong, Quoc~V Le, Ilya Sutskever, Oriol Vinyals, and Lukasz Kaiser.
  2016.
\newblock Multi-task sequence to sequence learning.
\newblock \emph{In Proceedings of ICLR 2016}.

\bibitem[{Luong et~al.(2015{\natexlab{a}})Luong, Pham, and
  Manning}]{luongeffective}
Minh-Thang Luong, Hieu Pham, and Christopher~D Manning. 2015{\natexlab{a}}.
\newblock Effective approaches to attention-based neural machine translation.
\newblock \emph{In Proceedings of EMNLP 2015}.

\bibitem[{Luong et~al.(2015{\natexlab{b}})Luong, Sutskever, Le, Vinyals, and
  Zaremba}]{luongaddressing}
Minh-Thang Luong, Ilya Sutskever, Quoc~V Le, Oriol Vinyals, and Wojciech
  Zaremba. 2015{\natexlab{b}}.
\newblock Addressing the rare word problem in neural machine translation.
\newblock \emph{In Proceedings of ACL 2015}.

\bibitem[{Meng et~al.(2016)Meng, Lu, Li, and Liu}]{meng2016interactive}
Fandong Meng, Zhengdong Lu, Hang Li, and Qun Liu. 2016.
\newblock Interactive attention for neural machine translation.
\newblock In \emph{In Proceedings of COLING 2016}.

\bibitem[{Mi et~al.(2016{\natexlab{a}})Mi, Wang, and
  Ittycheriah}]{mi2016supervised}
Haitao Mi, Zhiguo Wang, and Abe Ittycheriah. 2016{\natexlab{a}}.
\newblock Supervised attentions for neural machine translation.
\newblock \emph{In Proceedings of EMNLP 2016}.

\bibitem[{Mi et~al.(2016{\natexlab{b}})Mi, Wang, and
  Ittycheriah}]{Mi2016Vocabulary}
Haitao Mi, Zhiguo Wang, and Abe Ittycheriah. 2016{\natexlab{b}}.
\newblock Vocabulary manipulation for neural machine translation.
\newblock \emph{In Proceedings of ACL 2016}.

\bibitem[{Papineni et~al.(2002)Papineni, Roukos, Ward, and
  Zhu}]{papineni2002bleu}
Kishore Papineni, Salim Roukos, Todd Ward, and Wei-Jing Zhu. 2002.
\newblock Bleu: a method for automatic evaluation of machine translation.
\newblock \emph{In Proceedings of ACL 2002}, pages 311--318.

\bibitem[{Sennrich et~al.(2016{\natexlab{a}})Sennrich, Haddow, and
  Birch}]{sennrich2015improving}
Rico Sennrich, Barry Haddow, and Alexandra Birch. 2016{\natexlab{a}}.
\newblock Improving neural machine translation models with monolingual data.
\newblock \emph{In Proceedings of ACL 2016}.

\bibitem[{Sennrich et~al.(2016{\natexlab{b}})Sennrich, Haddow, and
  Birch}]{sennrich2015neural}
Rico Sennrich, Barry Haddow, and Alexandra Birch. 2016{\natexlab{b}}.
\newblock Neural machine translation of rare words with subword units.
\newblock \emph{In Proceedings of ACL 2016}.

\bibitem[{Shen et~al.(2015)Shen, Cheng, He, He, Wu, Sun, and
  Liu}]{shen2015minimum}
Shiqi Shen, Yong Cheng, Zhongjun He, Wei He, Hua Wu, Maosong Sun, and Yang Liu.
  2015.
\newblock Minimum risk training for neural machine translation.
\newblock \emph{In Proceedings of ACL 2015}.

\bibitem[{Sutskever et~al.(2014)Sutskever, Vinyals, and
  Le}]{sutskever2014sequence}
Ilya Sutskever, Oriol Vinyals, and Quoc~V Le. 2014.
\newblock Sequence to sequence learning with neural networks.
\newblock \emph{In Proceedings of NIPS 2014}.

\bibitem[{Tu et~al.(2016)Tu, Lu, Liu, Liu, and Li}]{tu2016coverage}
Zhaopeng Tu, Zhengdong Lu, Yang Liu, Xiaohua Liu, and Hang Li. 2016.
\newblock Coverage-based neural machine translation.
\newblock \emph{In Proceedings of ACL 2016}.

\bibitem[{Wang et~al.(2017)Wang, Lu, Tu, Li, Xiong, and Zhang}]{wang2017neural}
Xing Wang, Zhengdong Lu, Zhaopeng Tu, Hang Li, Deyi Xiong, and Min Zhang. 2017.
\newblock Neural machine translation advised by statistical machine
  translation.
\newblock \emph{In Proceedings of AAAI 2017}.

\bibitem[{Wu et~al.(2016)Wu, Schuster, Chen, Le, Norouzi, Macherey, Krikun,
  Cao, Gao, Macherey et~al.}]{wu2016google}
Yonghui Wu, Mike Schuster, Zhifeng Chen, Quoc~V Le, Mohammad Norouzi, Wolfgang
  Macherey, Maxim Krikun, Yuan Cao, Qin Gao, Klaus Macherey, et~al. 2016.
\newblock Google's neural machine translation system: Bridging the gap between
  human and machine translation.
\newblock \emph{arXiv preprint arXiv:1609.08144}.

\bibitem[{Zhai et~al.(2012)Zhai, Zhang, Zhou, Zong et~al.}]{zhai2012tree}
Feifei Zhai, Jiajun Zhang, Yu~Zhou, Chengqing Zong, et~al. 2012.
\newblock Tree-based translation without using parse trees.
\newblock \emph{In Proceedings of COLING 2012}.

\bibitem[{Zhang et~al.(2014)Zhang, Liu, Li, Zhou, and Zong}]{zhang:2014}
Jiajun Zhang, Shujie Liu, Mu~Li, Ming Zhou, and Chengqing Zong. 2014.
\newblock Bilingually-constrained phrase embeddings for machine translation.
\newblock \emph{In Proceedings of ACL 2014}.

\bibitem[{Zhang and Zong(2016)}]{zhang2016exploiting}
Jiajun Zhang and Chengqing Zong. 2016.
\newblock Exploiting source-side monolingual data in neural machine
  translation.
\newblock \emph{In Proceedings of EMNLP 2016}.

\bibitem[{Zhou et~al.(2017)Zhou, Hu, Zhang, and Zong}]{Zhou:2017}
Long Zhou, Wenpeng Hu, Jiajun Zhang, and Chengqing Zong. 2017.
\newblock Neural system combination for machine translation.
\newblock \emph{In Proceedings of ACL 2017}.

\end{thebibliography}
\bibliographystyle{ijcnlp2017}

\end{CJK*}
\end{document}